# Lexicon management and standard formats

## Éric Laporte

Institut Gaspard-Monge (IGM) - CNRS & University of Marne-la-Vallée
15, bd Descartes
F 77474 Marne-la-Vallée CEDEX 2 - France
eric.laporte@univ-mlv.fr

### Abstract

International standards for lexicon formats are in preparation. To a certain extent, the proposed formats converge with prior results of standardization projects. However, their adequacy for (i) lexicon management and (ii) lexicon-driven applications have been little debated in the past, nor are they as a part of the present standardization effort. We examine these issues. IGM has developed XML formats compatible with the emerging international standards, and we report experimental results on large-coverage lexica.

## Introduction

International standards for lexicon formats are in preparation, in order to facilitate associated software development, meta-documentation and exchange of language resources. To a certain extent, the proposed formats converge with prior results of standardization projects. However, their adequacy for (i) lexicon management and (ii) lexicon-driven applications have been little debated in the past, nor are they as a part of the present standardization effort. We examine these issues. IGM has developed XML formats of lexical resources compatible with the emerging international standards, and corresponding software tools, and carried out experimentation on large-coverage lexica of English and other languages. We report experimental results.

In the next section, we briefly describe the standard lexicon model presently in construction. Section 2 examines how adequate this model is for lexicon management. Section 3 takes into account the requirements of lexicon-based lexical tagging. The conclusion synthesizes our results.

## 1. Previous work

A series of standards of representation of lexica for natural language processing (NLP) were successively proposed, from Genelex (Normier, Nossin, 1990) to the present ISO group on Language resource management (Ide, Romary, 2002). Though some authors emphasize the differences between formats of lexica for written text processing currently in use (Wittenburg et al., 2002), there is much in common among the various models, which seem to be converging to an emerging ISO standard. IGM participates in this effort through the Outilex and Normalangue projects[1]. In this section, we describe the overall structure of the emerging standard and in particular we examine how it handles the dichotomy between lemma and inflected form in inflectional languages.

### 1.1. Lemmas

All proposed models have a lemma-based overall structure. In a lemma-based model, the set of lexical items is a set of lemma entries, i.e. nodes each of which represents a lemma of the language (Fig. 1).

```
<dic>
<entry>
 <lemma>game</lemma>
 <pos name='noun'/>
 <f name='reliability' value='1'/>
 <inflection>...
 </inflection>
</entry>
</dic>
```

Figure 1: Sample of a lemma-based lexicon

The notion of lemma exists in all languages. Part-of-speech, an essential feature, is attached to lemma entries. In the Olif model (Lieske et al., 2001), which integrates terminological with other lexical information and is consistent with international standards in terminography (TMF: Romary, 2001), terminological information is attached to lemmas. Higher-level features are attached either to lemma entries or to senses, which are themselves attached to lemmas but have a finer granularity. In the draft model of the Lexical Resource Markup Framework (LMF, ISO TC 37/SC4: Francopoulo 2003; George, 2003), features such as the applicability of syntactic constructions are attached to senses. Senses play the part of the nodes of a thesaurus; semantic links are attached to them. In the Papillon model (Boitet et al., 2002), multilingual links are attached to senses.

### 1.2. Inflection

The capacity to provide links between lemmas and inflected forms is part of the information contained in a lexicon of an inflectional language. Inflectional information in a lemma-based lexicon model can be specified in the form of inflectional rules, or of a complete paradigm of inflected forms attached to the lemma. In the second case, we obtain a variant of the lemma-based model, in which elements which represent inflected forms of a lexical item (word-form entries) are embedded in the corresponding lemma entry. We call this variant a mixed model, because it combines these two types of entries (Fig. 2). Inflectional features such as number, person, mood, tense, gender, case, etc. are attached to word-form entries.

---

[1] This paper owes much to the discussions inside this group, and to the anonymous reviewers' interesting remarks and constructive suggestions.

```
<dic>
<entry>
  <lemma>game</lemma>
  <pos name='noun'/>
  <f name='reliability' value='1'/>
  <inflected>
    <form>game</form>
    <f name='number' value='singular'/>
  </inflected>
  <inflected>
    <form>games</form>
    <f name='number' value='plural'/>
  </inflected>
</entry>
</dic>
```

Figure 2: Sample of a mixed lexicon

The Olif model has a lemma-based structure, but is adaptable to a mixed-base structure as a user extension. The draft LMF follows the same policy.

## 1.3. XML formats

IGM is responsible for the maintenance and management of the LADL's lexica, a set of large-coverage lexica of English and other languages with morpho-syntactic and syntactico-semantic information. In this work, we focused on morpho-syntax. A substantial part of these lexica is publicly available in plain form with the LGPL-LR license.

Since the 80s, the LADL's morpho-syntactic lexica have been managed in the Dela format (Courtois, 1990). In the recent years, more and more lexica and lexical databases have been proposed in XML formats, which provides several advantages: object-oriented data structures for software development can be automatically derived from XML structure (Hayashi, Hatton, 2001); XML encoding and associated grammars (DTDs or XML schemas) provide self-understandable meta-documentation about the content of resources; many languages and APIs are able to process XML resources; and XML formats are likely to be a good option for long-term storage.

An XML equivalent of the Dela format and a simple DTD were developed at IGM by Olivier Blanc in the framework of the Outilex project. Both this format and the underlying model of the Dela format are consistent with the morpho-syntactic part of the LMF draft model. As a major example of set of large-coverage lexica, the LADL's lexica were taken into account in several projects of lexicon standards, among them Genelex, whose results were largely integrated in later projects, and now LMF. Fig. 2 is a sample of this XML format. Meaningful names of features, like 'number', are attribute values, as in the draft ISO proposal about feature structure representation (Lee et al., 2004), and not element names, as in the Olif format. This choice makes it possible to write a less language-dependent DTD or XML schema, but the two solutions are largely equivalent.

As compared to the Dela format, the XML format tends to make lexical resources compatible with NLP software. The XML encoding of inflected forms can be used for word tags in annotated text, since it contains lexical information on the word. As Grover & Lascarides notice (2001), it is convenient to represent lexical features in the same way in lexica and in annotated text, as is done in the Intex (Silberztein, 1994) and Unitex (Paumier, 2002) systems. In addition, the use of tags which have an internal structure is an improvement upon the most widely used annotated corpora and their unstructured word tags, now obsolete for natural language processing. The possibility of tags for multi-word units is also an improvement upon the notion of word in the TEI (Ide, Véronis, 1995), since this notion misses the essential distinction between graphically delimited tokens and words in the linguistic sense (Gross, 1986).

Finally, a straightforward generalization of our XML encoding of inflected forms is used for lexical masks. A lexical mask (Blanc, Dister, 2004) is a request to search an annotated text for a word. It specifies a word or a set of words to be matched in the text. A lexical mask is a possibly underspecified word tag. A sequence of lexical masks, or a finite automaton over lexical masks, defines a pattern to be matched in annotated text.

Software tools for text annotation and searching have been implemented on the basis of this XML format and evaluated in the framework of the Outilex project.

## 2. Adequacy for lexicon management

Lexicon management is one of the activities for which the emerging ISO model should be adequate. In this section, we introduce the notion of lexicon management, we explain why it is relevant for NLP and we survey current practices. We examine how adequate the model and its variants are for lexicon management. We report experimentation with a large-coverage lexicon.

### 2.1. Lexicon management

Reusing existing lexica provides an obvious means of alleviating the initial effort for the development of new applications. The construction of standard exchange data formats already improved the reusability of lexica. However, reusing a lexicon implies it should have a certain degree of flexibility. A lexicon is not a static resource, it evolves with time. Due to the evolution of language across time, and especially of technical language, regular updates are necessary; a new application of a lexicon may involve the selection of a domain-specific vocabulary. Lexicon reuse is likely to be facilitated by the availability of tools and data for maintenance and adaptation of the resources, and more generally for language resource management.

Standards are usually inspired from current practices. Examples of systems with language resource management tools are Xelda (Poirier, 1999), Intex and Unitex. However, most other general-purpose systems for language resource processing and text processing are deprived of even basic functionality for lexicon management. The design of the Gate system (Cunningham, 2002), for example, ensures that the user never performs any operation of language resource management, since every resource used has to be wrapped in a text-processing tool (*ibid.*, p. 228). With a British sense of understatement, Cunningham (2002, p. 249) admits that Gate neglects data resource components. The open-source, general-purpose computational-linguistics toolkit NLTK (Loper, Bird, 2002; Bird, Loper, 2004) does not offer functionality for resource management either.

In this context, few language engineering companies are aware of modern lexical resource management, and the NLP research community pays insufficient attention to issues such as the improvement of current techniques and their extension to new languages. Merely taking into account current practices would hardly ensure that emerging ISO standards will facilitate resource management practices and help designers and implementers of applications to achieve a better integration between language resource management and their other activities. Rather, the issue deserves special attention.

## 2.2. Adequacy of the lemma-based model

The basic elements of the model, i.e. lemma and sense, have the levels of granularity required for attaching the lexical properties with limited redundancy: morphosyntactic information to lemmas; syntactico-semantic, terminological and multilingual information to senses. In the case of inflectional languages, information related to inflectional morphology is attached to lemmas. However, recall (1.2) that current versions of the model do not define whether this information should take the form of compact inflectional rules, or of a set of word-form entries. The latter case defines the mixed model. Has this alternative any consequences on lexicon management?

The cost and error rate of update operations are reduced when they are performed on compact resources with limited redundancy. A set of word-form entries is more redundant than a representation of an inflectional pattern. Firstly, the word stem is duplicated several times in word-form entries, whereas it is not included in inflectional rules. The ratio between the number of lemmas and of word-forms has an order of magnitude of 10 in inflectional languages. Secondly, the representation of an inflectional pattern can be shared by several lemma entries. The ratio between the number of inflectional patterns and of lemmas has an order of magnitude of 1 000, multi-word units excluded.

In addition, according to specialists, interactive construction and update of lexica is facilitated and less errorprone when independent entries are displayed readably and with a large density of information, e.g. with several dozens of entries and several properties in the same page or window. Due to redundancy, such an edition-oriented viewer is simpler to implement if the data follow the lemma-based, rather than the mixed, model.

Finally, the lemma-based model is more general than the mixed one, which is specific to inflectional languages. The lemma-based model is adapted to languages without inflection, such as Chinese, by deleting inflectional information. It is also adequate for agglutinative languages, such as Turkish, in which a word is a sequence of graphically non-delimited morphemes. For such languages, information about the compatibility of the lemma (and its morphological variants) with grammatical morphemes (e.g. Oflazer, Inkelas, 2003) must be substituted to inflectional information. The mixed model is not applicable to these languages because the set of word-forms generated from a lemma is large or infinite.

A third type of model exists: the wordform-based model, with all lexical information attached to elements which represent inflected forms of lexical items (Fig. 3). ELRA's catalogue of written-text processing lexica contains several word-form lexica. A number of NLP systems use word-form lexica, but usually not for lexicon management. The word-form model is adapted only to a corpus-based approach of lexicon management. For example, if a corpus representative of a domain is used to extract a sub-lexicon from an existing lexicon, extraction can easily be performed at the level of word-forms: only those words that occur in the corpus are retained. Similarly, a word-form lexicon can be extended on the basis of a corpus by finding new inflected forms in texts, guessing at lexical information (cf. Quasthoff, 1998) and inserting them directly. However, this approach has the same drawbacks as the mixed model, plus a serious one: it is inefficient in terms of lexicon management, since when an inflected form is inserted, the other inflected forms of the same lemma are not.

```
<dic>
  <entry>
    <form>game</form>
    <lemma>game</lemma>
    <pos name='noun'/>
    <f name='reliability' value='1'/>
    <f name='number' value='singular'/>
  </entry>
  <entry>
    <form>games</form>
    <lemma>game</lemma>
    <pos name='noun'/>
    <f name='reliability' value='1'/>
    <f name='number' value='plural'/>
  </entry>
</dic>
```

Figure 3: Sample of a wordform-based lexicon

What is the role of the mixed model for inflectional languages? Though not applicable to agglutinative languages and not best adapted to lexicon updating operations, it can be used as an intermediary during the conversion of a lemma-based to a wordform-based lexicon. Conversion between XML formats is relatively easy and efficient. From a lemma-based lexicon, a mixed lexicon is obtained by applying inflectional rules and generating inflected forms; from a mixed-model lexicon, a word-form lexicon is obtained by copying lemma entries into the corresponding word-form entries. In addition, the mixed model is adequate for storage and exchange by organizations who do not practice lexicon management, or do not own lexicon management resources and tools such as inflectional patterns and generators of inflected forms. As a matter of fact, converting a mixed-model lexicon to a word-form lexicon does not involve such technology.

## 2.3. Tests

Models must be tested in order to go beyond the status of administrative recommendations. The objectives of our tests were to implement lossless converters between the Dela format and the XML format (1.3), and to know the typical size of an XML mixed-model lexicon. The Dela format is validated by a 20-year tradition of large-coverage lexicon management. For the tests, we used the LADL's lexica (1.3).

Among the Dela formats, the closest equivalent to the mixed model is the Delaf format for word-form lexica. A converter from Delaf to XML in C++ (O. Blanc) and a converter from XML to Delaf in XSLT have been implemented. Large-coverage inflected-form lexica of English and French have been converted in both directions. A 48-Mb Delaf lexicon of 122 000 lemmas and 1 260 000 word forms is converted into a 262-Mb XML lexicon in 2 mn 22 s. The reverse conversion took 36 mn. These tests show that the size of an XML mixed-model lexicon and conversion times are both practicable.

## 3. Adequacy for lexicon-based analysis

Examples of lexicon-based applications on written text are: text generation, text correction, text search, translation, and corpus annotation. Lexicon-based NLP involves lexical tagging. Standard models and structures for lexica should not create obstacles to this essential operation. In this section, we select the most accurate and efficient technique for lexicon-based analysis, we assess the corresponding lexicon structure as a potential standard model, and we report the results of our tests about format conversion and quick lookup.

### 3.1. Approaches to lexicon-based tagging

The mixed model and the word-form model are adapted to text processing, since forms occurring in texts are directly described in lexical entries and these descriptions are accessible through efficient lookup. The only variation to be taken into account at lookup time between forms occurring in texts and forms described in the lexicon is uppercase vs. lowercase. The lemma-based model can also be used in applications, but this requires choosing one of the following two solutions:
- lemmatizing the text by applying a tagger or a stemmer (Porter, 1980), which gives only approximate results;
- access the lexicon through morphological analysis (Lezius, 2000), a computation much more complex than that required by uppercase vs. lowercase variation, which slows down lexicon lookup, a highly repetitive operation. The only way to combine the two constraints (accurate results and computational efficiency) is to manage jointly a lexicon of lemmas and a word-form or mixed-model lexicon. The classical solution for this is to compile the first into the latter by automated inflection, i.e. generation of inflected forms from lemmas (e.g. Domenig, 1988; Courtois, 1990).

Automated inflection and word-form lexicon lookup are therefore basic techniques, but they are supported only by a few systems like Intex and Unitex, and little referred to by scientific literature. Even the most general articles about lexica discuss issues such as the internal structure of entries, the role of rules in the construction and interpretation of a lexicon (Briscoe, 1991), or multiple views of underlying lexicon entries (Gibbon, Trippel, 2000), but are silent about issues connected to automated inflection. So are entire general-purpose NLP handbooks like Jurafsky, Martin (2000)[2]. Many researchers and engineers are content with statistical taggers or stemmers. However, these approaches are only approximations, and language engineering resorts to them in the absence of resources of sufficient quality and coverage. If the sole approximate approaches were taken into account for the design of language resource standards, we would be taking for granted that the content of the resources is deficient. On the contrary, the advancement of technologies connected to word-form lexica is a factor of progress in NLP, and the structure of the corresponding resources deserves special attention.

### 3.2. Index-based lexicon models

The overall structures adapted to efficient word-form lexicon lookup are index-based structures such as a tree or finite-state automaton structure, and can be specific to a database management system. With these structures, lookup time depends on the size of the text, but not on the number of entries in the lexicon. However, the underlying model of index-based structures is very different from the entry-based models mentioned so far. The notions of entry (lemma, sense or word form) are subordinated to tree nodes, automaton states or transitions. The described words are scattered in nodes or transitions.

In a context where activities related to text processing applications are much more frequent than language resource management, as in the Gate and NLTK views of NLP, index-based lexicon models could be seen as candidates for standard exchange models. However, such models would have two serious drawbacks.
- They are dependent on methods and tools of lexicon compression and quick lookup, which are numerous and likely to undergo further variations (Laporte, 2005).
- They are incompatible with interactive update. Therefore, using them as exchange models would cause a separation between lexicon management and lexicon-based text processing. We should rather consider these two activities as interdependent. For example, the behaviour of a system depends on the selection of the resources that support its operation; and feedback of performances into resource contents is a principled approach to the control of the improvement of a system.

The use of the lemma-based or mixed models as exchange models does not present the same drawbacks. In inflectional languages, compiling a mixed-model or word-form lexicon into an index-based lexicon is the standard means of updating the latter, and an efficient, essential resource management operation.

### 3.3. Tests

The objectives of our tests were to check the compatibility of the mixed model with methods of lexicon compression and quick lookup. We used the same XML format as above, and the LADL's methods of large-coverage lexicon compression, validated by a 15-year tradition. A converter from XML mixed-model lexica to an index-based format has been implemented in C++ (O. Blanc, 2003). The XML lexicon is read by the XmlTextReader API. The index-based format is a finite-state structure. The 262-Mb lexicon of 122 000 lemmas and 1 260 000 word forms is converted in 1 mn 22 s. The index-based lexicon has 71 000 states and 160 000 transitions and a size of 9.6 Mb, (3.7 % of the original

---

XML file). The same XML file compressed by gzip has a size of 6.1-Mb file (2.3 %).

Access to the index-based lexicon has been implemented in C++. The lexicon is searched for all the words of a corpus of 1 million words in 1 mn.

These tests show that the file sizes and computation times involved in the compilation and use of an index-based lexicon are compatible with the emerging lexicon models and with their XML implementation.

## Conclusion

We have assessed the validity of emerging ISO standards for lexicon models and formats. We examined two issues little debated in past and present literature: lexicon management and lexicon-based tagging. The lemma-based model appears as adequate for lexicon management, provided that practical tools for lexicon management are available, viz. an edition-oriented viewer and a generator of inflected forms. For inflectional languages, the mixed model is an alternative solution for storage and exchange. Our tests with large-coverage lexica validate both models.

## References

Bird, S., E. Loper (2004). "NLTK: the Natural Language Toolkit". In *Proc. of ACL*.

Blanc, O. (2003). *Rapport d'avancement Outilex*, IGM.

Blanc, O., A. Dister (2004). "Automates lexicaux avec structure de traits". In *RECITAL 2004, Fès*, pp. 23-32.

Boitet, Ch., M. Mangeot, G. Sérasset (2002). "The PAPILLON project: cooperatively building a multilingual lexical data-base to derive open source dictionaries & lexicons". In *COLING Workshop on NLP and XML*, Taipei, Taiwan, pp. 93-96.

Briscoe, T. (1991). "Lexical issues in Natural Language Processing". In E. Klein and F. Veltman, eds., *Natural Language and Speech*, Springer.

Courtois, B. (1990). "Un système de dictionnaires électroniques pour les mots simples du français", *Langue Française* 87, Paris: Larousse.

Cunningham, H. (2002). "GATE, a general architecture for text engineering". *Computers and the Humanities* 36:223-254.

Domenig, M. (1988). "Word Manager: A System for the Definition, Access and Maintenance of Lexical Databases". In *Proc. of Coling*, Budapest, Vol. 1.

Francopoulo, G. (2003). *Proposition de norme des lexiques pour le traitement automatique du langage*, AFNOR, 21 p.

George, M. (2003). *Terminology and other language resources. Lexical Resource Markup Framework*, ISO, 16 p.

Gibbon, D., Th. Trippel (2000). "A multi-view hyper-lexicon resource for speech and language system development". *Proc. of LREC*, Athens, pp. 1713-1718.

Gross, M. (1986). "Lexicon-Grammar. The Representation of Compound Words". In *Proc. of COLING*, Bonn, pp. 1-6.

Grover, C., A. Lascarides (2001). "XML-Based Data Preparation for Robust Deep Parsing". In *Proc. of the Joint EACL-ACL Meeting*, Toulouse.

Hayashi, L., J. Hatton (2001). "Combining UML, XML and relational database technologies. The best of all worlds for robust linguistic databases". In *Proc. of the IRCS Workshop on Linguistic Databases*.

Ide, N., L. Romary (2002). "Standards for language resources". *Proc. of LREC*, pp. 839-844, Las Palmas.

Ide, N., J. Véronis (1995). *Text Encoding Initiative: Background and Context*. Dordrecht: Kluwer.

Jurafsky, D., J. Martin (2000). *Speech and language processing*. Prentice Hall. 934 p.

Laporte, É. (2005). "Symbolic natural language processing". In *Applied Combinatorics on Words*, Lothaire, Cambridge Univ. Press, pp. 153-195.

Lee, Kiyong, H. Bunt, S. Bauman, L. Burnard, L. Clément, É. de la Clergerie, Th. Declerck, L. Romary, A. Roussanaly, C. Roux (2004). "Towards an international standard on feature structure representation". In *Proc. of LREC*, pp. 373-376.

Lezius, W. (2000). "Morphy. German Morphology, Part-of-Speech Tagging and Applications", in *Proc. of EURALEX*, pp. 619-623, Stuttgart.

Lieske, Ch., S. McCormick, G. Thurmair (2001). "The Open Lexicon Interchange Format (OLIF) Comes of Age". *Machine Translation Summit VIII*.

Loper, E., S. Bird (2002). "NLTK: the Natural Language Toolkit". In *Proc. of the ACL Workshop on Effective Tools and Methodologies for Teaching Natural Language Processing and Computational Linguistics*, Philadelphia.

Normier, B., M. Nossin (1990). "GENELEX Project: Eureka for Linguistic Engineering", *Proc. of the International Workshop on Electronic Dictionaries*, OISA, Kanagawa, Japan, pp. 63-70.

Oflazer, K., Sh. Inkelas (2003). "A Finite State Pronunciation Lexicon for Turkish". In *Proceedings of the EACL Workshop on Finite State Methods in NLP*, Budapest.

Paumier, S. (2002). *Unitex. Manuel d'utilisation*, Research report.

Poirier, H. (1999). *The XELDA framework*. http://www.dcs.shef.ac.uk/~hamish/dalr/baslow/xelda.pdf

Porter, M.F. (1980). "An algorithm for suffix stripping". *Program* 14(3):130-137.

Quasthoff, U. (1998). "Tools for Automatic Lexicon Maintenance: Acquisition, Error Correction, and the Generation of Missing Values". In *Proc. of LREC*, pp. 853-856.

Romary, L. (2001). *Towards an Abstract Representation of Terminological Data Collections. The TMF model*. TAMA. Antwerp.

Silberztein, M. (1994). "INTEX: a corpus processing system". In *Proc. of COLING*, Kyoto.

Wittenburg, P., W. Peters, S. Drude (2002). "Analysis of Lexical Structures from Field Linguistics and Language Engineering". In *Proc. of LREC*.